\documentclass[]{spie}  
\usepackage[]{graphicx}
\usepackage{multirow}

\title{End-to-End Defect Detection in Automated Fiber Placement Based on Artificially Generated Data}

\author{Sebastian Zambal\supit{a} and Christoph Heindl\supit{a} and Christian Eitzinger\supit{a} and Josef Scharinger\supit{b}
\skiplinehalf
\supit{a}Profactor GmbH, Im Stadtgut A2, 4407 Steyr, Austria\\
\supit{a}JKU Institute of Computational Perception, Altenbergerstr. 69, 4040 Linz, Austria\\
}

\authorinfo{Contact: Sebastian Zambal (sebastian.zambal@profactor.at)}

\begin{document} 
  \maketitle 

\begin{abstract}
Automated fiber placement (AFP) is an advanced manufacturing technology that increases the rate of production of composite materials. At the same time, the need for adaptable and fast inline control methods of such parts raises. Existing inspection systems make use of handcrafted filter chains and feature detectors, tuned for a specific measurement methods by domain experts. These methods hardly scale to new defects or different measurement devices. In this paper, we propose to formulate AFP defect detection as an image segmentation problem that can be solved in an end-to-end fashion using artificially generated training data. We employ a probabilistic graphical model to generate training images and annotations. We then train a deep neural network based on recent architectures designed for image segmentation. This leads to an appealing method that scales well with new defect types and measurement devices and requires little real world data for training.
\end{abstract}


\keywords{AFP, deep learning, probabilistic graphical model, segmentation, end-to-end}

\begin{figure}[h]
\begin{center}
\begin{tabular}{c}
\includegraphics[width=0.99\textwidth]{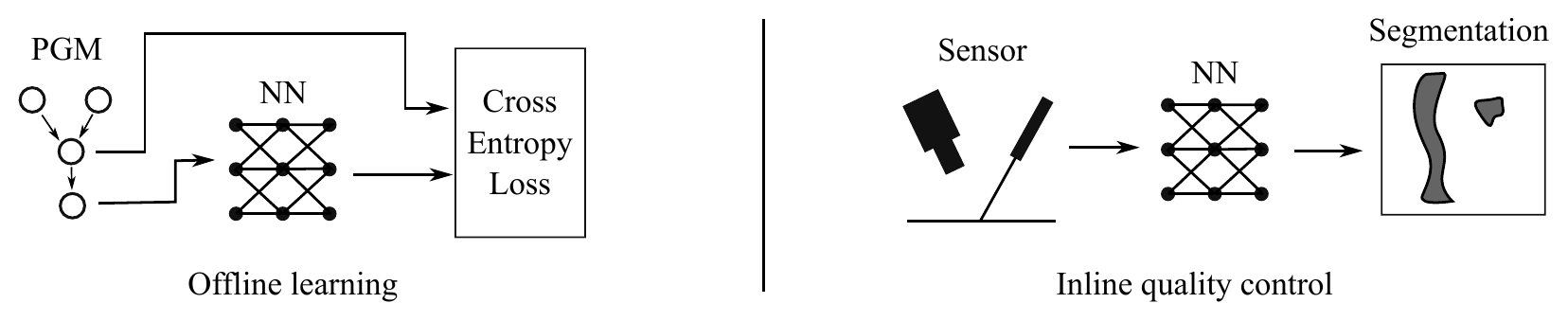}
\end{tabular}
\end{center}
\caption[Overview]
{ 
\label{fig:overview}: Overview of our method: During the offline learning process, a probabilistic (graphical) model generates random domain data, that is used for training an image-to-image neural network on a segmentation task (left). For monitoring during AFP, the trained neural network is used for segmentation of defects.
}
\end{figure} 

\section{INTRODUCTION}
\label{sec:intro}  

Recent advances in deep learning brought great improvements in solving many hard computer vision tasks. For many applications, databases with large amounts of data are available for training of deep neural networks (e.g. the KITTI Vision Benchmark Suite \cite{Geiger2012} for autonomous driving). For many industrial quality control applications, however, it is very hard to collect large amounts of data – at least in early design stages before hardware installation. Images that are acquired in quality control vision applications typically depict very application-specific objects and often require some expert knowledge for interpretation. Labeling large amount of data is costly and often fails due to the lack of domain experts. Additionally, specific defects may occur very rarely.

The idea of using artificial training data has been addressed by several authors in the past. Using 3D renderings of street scenes based on a conventional gaming engine was proposed for autonomous driving \cite{johnson2016driving}. To perform 3D hand tracking based on 2D images, artificially generated data was used \cite{Mueller2018}. Heindl et al. \cite{heindlrobotpose} predict robot poses in photos from generated data.

In this paper, we propose a novel structured approach for the use of artificial data to enable the application of machine learning methods for a specific quality control vision application: Monitoring of automated fiber placement (AFP) processes. The vision system involved is a laser triangulation sensor that is mounted on the lay-up machinery. This  sensor delivers depth information immediately after placement of carbon fiber tows. We employ a probabilistic graphical model to describe the creation of depth maps. Expert knowledge and process parameters can easily be included into the model. Depth maps sampled from the model are used to train a deep neural network to perform the actual task of image segmentation. A real data example is used to assess segmentation performance of the deep neural network that was exclusively trained on artificial data.  


\section{RELATED WORK}

\textbf{Data generation} The lack of annotated datasets for supervised machine learning has begun to impede the advance of successful usage of such methods in industrial applications. To cope with this problem, a variety of methods have been proposed. Active learning methods \cite{druck2009active, settles2012active} better involve domain experts by presenting only those data samples of high value to the current training progress. Semi-supervised learning \cite{chapelle2009semi} transfers annotations from small labeled datasets to larger unlabeled ones by making task-specific smoothness assumptions. Weakly supervised learning \cite{zhou2017brief} attempts to infer precise labels from noisier, more global annotations, which are often easier to obtain. Transfer learning \cite{pan2010survey} limits the amount of required training data by adapting pre-trained models to specific tasks. 

In contrast, methods that make use of artificial data generation utilize a simulation engine that generates data along with ground truth labels. Such methods are gaining popularity \cite{johnson2016driving, peng2015learning, heindlrobotpose}, due to the availability of general purpose simulation engines. Like in this work, the simulator is driven by samples from a probabilistic model that has to be designed specific for the task at hands.

\textbf{Segmentation} Image segmentation is the task of partitioning an image into regions of common characteristics. Early works include image thresholding \cite{davis1975region} and clustering \cite{jain2010data}. With the raise of Convolutional Neural Nets (CNNs), image segmentation \cite{chen2014semantic, milletari2016v} is understood as image-to-image conversion. U-Nets \cite{ronneberger2015u}, which this work builds upon, consist of contracting and expanding data paths. This assures that global context and local details are both exploited to calculate the final segmentation result.

In contrast to the present work, quality inspection of automated fiber placement (AFP) is rarely formalized as segmentation problem that is learned end-to-end. Frequently, pipelines consisting of handcrafted filters and feature detectors are proposed, which are tuned for specific measurement devices. Cemenska et al. \cite{cemenska2015automated} proposes automatic detection of ply boundaries and tow ends based on laser profilometers. Juarez et al. \cite{juarez2016advances} studies the usage of thermographic cameras for gap detection based on heat diffusion. Similarly, Denkena et al. \cite{denkena2016thermographic} use thermographic inspection to detect overlaps, gaps, twisted tows and bridges based on thresholding.

\section{QUALITY CONTROL FOR AUTOMATED FIBER PLACEMENT}

For production of carbon fiber reinforced plastics (CFRP) parts, typically layers of carbon fiber material are placed one layer after the other on some tooling. In automated fiber placement (AFP) this is done by large machines that are able to automatically lay-up tows (“stripes” of carbon fiber material). Typically, these machines are able to lay-up 8, 16, or even 32 tows in parallel next to each other. In order to assure mechanical stability of the final parts, it must be verified that carbon fibers are placed in the right way. There is a set of possible defects that relate either to incorrect placement of tows or foreign objects. 

While our method generalizes to any measurement method, we propose the use of a laser triangulation sensor to perform inline quality control. The type of data acquired by such a system is a sequence of laser line profiles. Our system accumulates multiple such profiles into a depth map where individual pixels describe depth. We are not so much interested in absolute depth values. Rather, we focus on small depth variations that reveal different defects on the surface. The specific surface defects that we address here are:
\begin{itemize}
\item{Gaps: Irregular larger spacings between neighboring tows.}
\item{Overlaps: Irregular overlaps of tows that should actually be placed next to each other.}
\item{Fuzzballs: Accumulation of carbon fibers that form small balls and fall onto the surface during lay-up.}
\end{itemize}
Besides the above defect types, we add regular tow surface to the list of possible class labels for the segmentation problem at hand. In order to perform quality control, it is necessary to assign one of these class labels to each pixel in the input depth map. Therefore, the problem that needs to be solved is classical image segmentation. 

\section{PROBABILISTIC MODEL}

Bayesian Networks are a concept to enable modeling of complex joint probability distributions in terms of usually sparser conditional probabilities. The modeling process typically starts with setting up a list of relevant entities of the problem. In the next step, conditional probability distributions are defined to model the relationships between the different entities. For both steps, expert knowledge is exploited. 

\begin{figure}
\begin{center}
\begin{tabular}{c}
\includegraphics[width=0.6\textwidth]{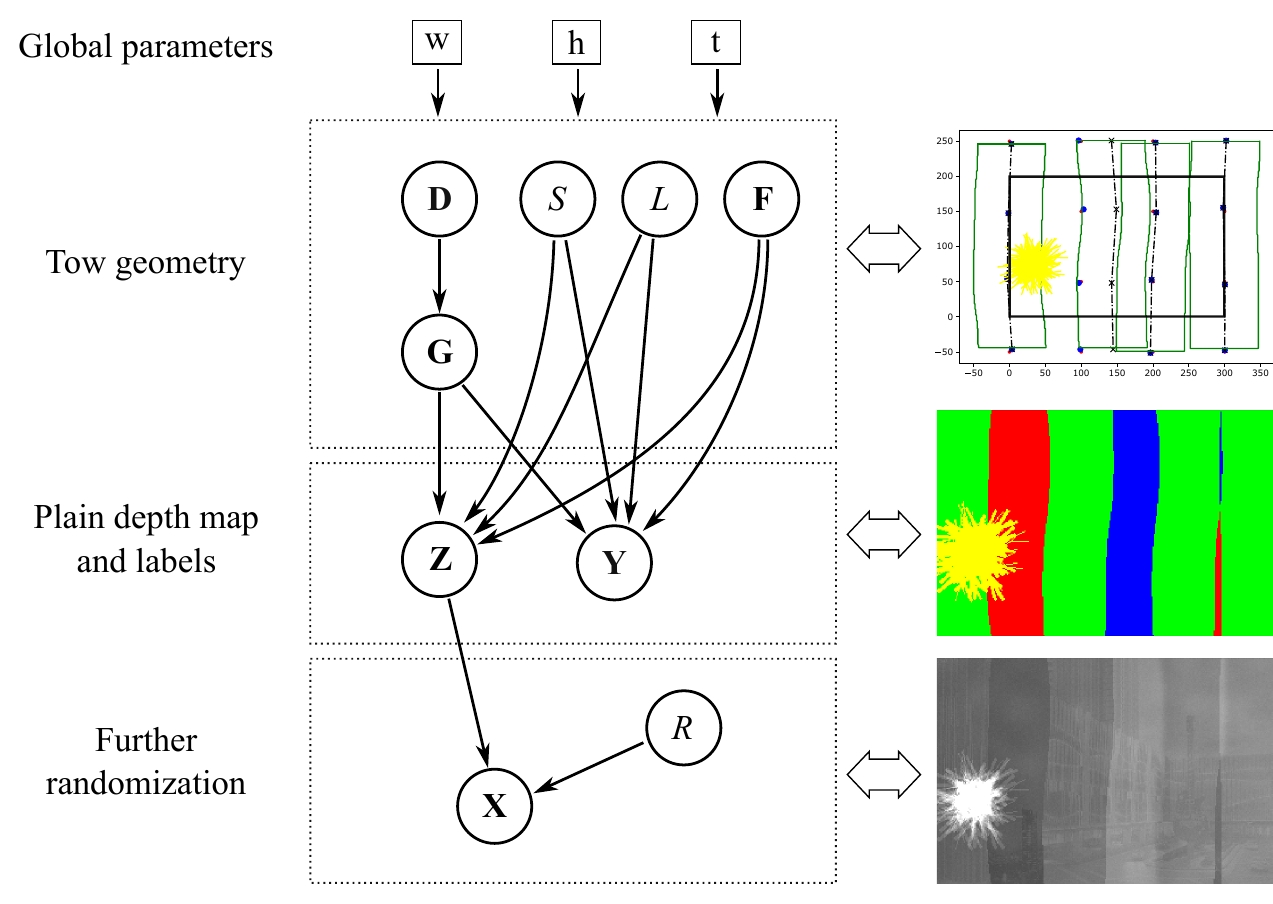}
\end{tabular}
\end{center}
\caption[probModel]
{ \label{fig:probModel}: Illustration of the probabilistic model. Random variables in different abstraction levels represent either high-level properties, labels (used as target output of neural network), and actual images (used as input for neural network). Pixel labels are visualized with different colors: gap (red), regular tow (green), overlap (blue), and fuzzball (yellow).}
\end{figure}


Figure \ref{fig:probModel} illustrates the structure of the Bayesian Network. There are four layers that make up our probabilistic model. Each layer builds upon the previous one. The last layer directly outputs artificial depth maps. In the following, we outline details about each of the layers.

\textbf{Global parameters:} At the top level, we model very basic properties of the image generation process: Size of the depth map $w \times h$ and width of individual tows $t$. The tow width is specified in terms of pixels. We use fixed values for $w$, $h$, and $t$ for experiments in this work. However, it would be possible to assign a probability distribution to the tow width in order to cover different sensor setups with varying field of view, resolution, or tow width.

\begin{figure}[h]
\begin{center}
\begin{tabular}{c}
\includegraphics[width=0.99\textwidth]{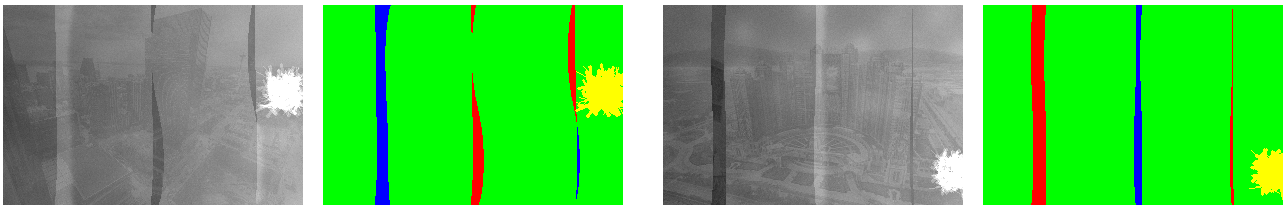}
\end{tabular}
\end{center}
\caption[trainingExamples]
{ 
\label{fig:trainingExamples}: Examples of artificial data: Depth maps (gray scale) and labels(colored). Pixel labels are visualized with different colors: gap (red), regular tow (green), overlap (blue), and fuzzball (yellow).
}
\end{figure} 

\textbf{Tow geometry} Based on global parameters, a rectilinear grid of control points is generated. This models the arrangement of individual tows in the simulated field of view. Each column of the grid corresponds to a single tow within the field of view. Initially, the spacing between control points is equal to tow width $t$. In order to model small deviations of tows from a perfect rectilinear grid, we add random displacements of control points. The size of displacements $\mathbf{D}$ (vector with displacements in x- and y-axis for all grid points) is assumed normally distributed with zero mean and standard deviation of 3\% of the tow width $t$. Besides small deviations, we are interested in modeling gaps and overlaps as they might occur in a real production environment. Therefore, we randomly select a single column in the control grid and apply a larger horizontal shift $S$ to its points. We assume $S$ to be uniformly distributed in the range $[0.05t, 0.5t]$. We denote the final vector of control grid coordinates by $\mathbf{G}$.

Besides gaps and overlaps, we intend to detect fuzzballs which may occur practically at any location within the field of view. The fuzzball center $\mathbf{F}$ is assumed uniformly distributed across the depth map. We model the geometry of a fuzzball simply by a set of individual thin fibers at random locations near the center of the fuzzball. The first end-point of the fiber is chosen near the fuzzball center. The second end-point is calculated by adding a vector with orientation uniformly distributed between $[-\pi, \pi]$. The length $L$ of the vector is uniformly distributed between $[v, 2v]$, where we choose a fixed value of 30 pixels for $v$.

\textbf{Plain depth map and labels} The next layer of the probabilistic model takes the geometric representation of tows and fuzzball and converts it into a depth map representation. To accomplish this, individual tow control points are converted to polygonal contours of tows. The resulting polygons are filled with a specific depth value to make tows more elevated than the background. This is accomplished by a standard polygon filling algorithm. In case of overlaps, the sequence in which the overlapping tows are added is important. In real data, an overlap has a sharp edge at the tow boundary of the top tow. At the edge of the lower tow, that is "buried" under the top tow, a smooth depth transition is typically observed. We account for this by explicitly applying a distance transform on one side of the overlap. We apply a sigmoid function across a fixed distance range to account for the smooth depth transition.

To derive the depth map $\mathbf{Z}$, the fuzzball is added at its sampled location. This is done by simply accumulating the individual simulated fibers as thin lines with fixed thickness. In addition to $\mathbf{Z}$, a map of pixel labels $\mathbf{Y}$ is calculated. $\mathbf{Y}$ serves as ground truth for neural network training.

\textbf{Further randomization} Real sensor data typically contains global geometry variations and exhibits some kind of texture. Our model accounts for both effects by additional randomized modifications of the depth map. The modifications are first calculated as separate depth maps. These are blended over the plain depth map $\mathbf{Z}$ to generate the final artificial depth map $\mathbf{X}$.

In order to account for global (low frequency) surface variation, we calculate a linear ramp with a slope in horizontal direction of the depth map. The ramp is zero at the horizontal center of the depth map and has a random slope $R$ which is assumed normally distributed with zero mean.

It is difficult to create a probabilistic model that generates a rich set of textures similar to those observed in real data. We avoid explicit modeling of such textures. Instead, we take an image database of 232 photos of urban scenes which we find convey similar image frequencies compared to real data. We first convert these images to gray scale. Then, each depth map is blended with two randomly chosen gray-scale photos: the first image is blended over regions of the top layer of tows, the second image is blended over regions of the bottom layer. 

By adding the above modifications, we force the subsequent neural network training to focus on the relevant content (signal) and ignore global depth variations, texture, or noise. Two examples of generated training samples (depth map and labels) are shown in figure \ref{fig:trainingExamples}.

\section{INFERENCE VIA NEURAL NETWORK}

It is easily possible to draw samples from the joint probability distribution $\mathrm{P}(S, \mathbf{F}, ... \mathbf{X}, \mathbf{Y}, \mathbf{Z})$ of the above outlined probabilistic model. 
For quality control, we are interested in the conditional probability (or maximum a-posteriori assignments) of labels $\mathbf{Y}$ given the observed depth map $\mathbf{X}$: $\mathrm{P}(\mathbf{Y} \mid \mathbf{X})$. We use a neural network to learn a distribution $\mathrm{Q}(\mathbf{Y} \mid \mathbf{X})$ that seeks to approximate this term. 

We deploy a neural network with an architecture similar to U-Nets \cite{ronneberger2015u}. U-Nets are instances of a broader class of so called pixel-to-pixel networks. In the context of segmentations, U-Nets are fed an input image that is transformed into an output image corresponding to the segmentation of the input. The input image undergoes a sequence of down-sampling and up-sampling steps. In addition, activations are forwarded from layers before down-sampling to layers after up-sampling. Down-sampling and up-sampling supports use of context information. Forwarding maintains spatial resolution.

\begin{figure}
\begin{center}
\begin{tabular}{c}
\includegraphics[width=0.99\textwidth]{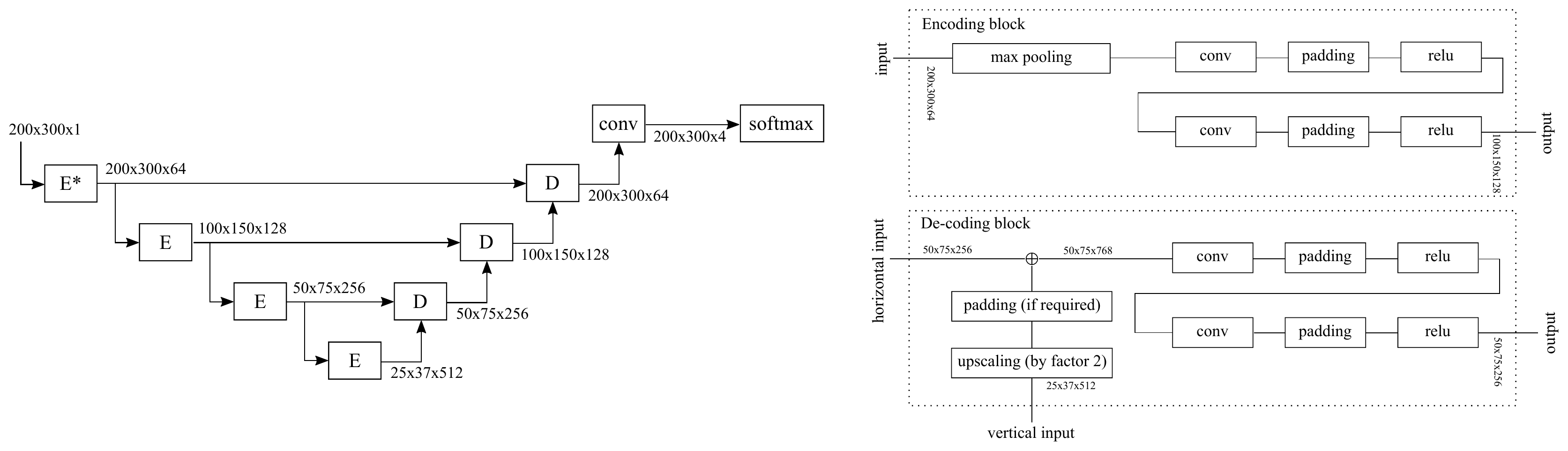}
\end{tabular}
\end{center}
\caption[trainingExamples]
{ 
\label{fig:model}: Network architecture used for segmentation involves encoding (E) and decoding (D) blocks. In the first encoding block E$^*$ the max pooling operation is skipped. The right part of the figure shows the internal structure of encoding and decoding blocks. The size of activations (height $\times$ width $\times$ features) shown in the figure are examples. See text for details.
}
\end{figure}

Our model is illustrated in figure \ref{fig:model}. The main building blocks are encoding blocks and decoding blocks. The internal structure of both is shown on the right in figure \ref{fig:model}. In the first encoding block E$^*$ the max pooling operation is skipped. The specified dimensions of tensors provided in the figure are only examples. In general, width and height of activations in each layer are half of those of the layer above. The number of features in each layer are twice of those of the layer above. Our model contains some modifications compared to the original U-Net architecture. Instead of transposed convolution we use simple spatial up-scaling in the decoding blocks. We avoid spatial shrinkage by introducing padding steps after convolution and after up-scaling. Therefore, we do not need cropping of activations that are forwarded from encoding blocks to decoding blocks at the same level.

The output of our model has the same width and height as the input image. We convert class scores per pixel to probability distributions over classes using the soft-max operator. During training we optimizes the pixel-wise cross-entropy between the true distribution $\mathrm{P}(Y_{i,j} \mid \mathbf{X})$ and its approximation $\mathrm{Q}(Y_{i,j}\mid \mathbf{X})$ given by 
\begin{equation}
  H(\mathrm{P},\mathrm{Q})=H(\mathrm{P})+D_{\mathrm{KL}}(\mathrm{P}||\mathrm{Q})  
\end{equation}
where $H$ denotes the (cross-)entropy and $D_\mathrm{KL}$ is the Kullback–Leibler (KL) divergence. Hence, training our neural network is equivalent to optimizing the KL divergence of $D_{\mathrm{KL}}\left[\mathrm{P}(\mathbf{Y} \mid \mathbf{X})||\mathrm{Q}(\mathbf{Y} \mid \mathbf{X})\right]$ where the distribution $\mathrm{Q}$ factors into 
\begin{equation}
\mathrm{Q}(\mathbf{Y} \mid \mathbf{X}) = \prod\limits_{i,j \in \Omega}\mathrm{Q}(Y_{i,j}\mid \mathbf{X})
\end{equation}
where $\Omega$ is the image domain.


\section{RESULTS}

\begin{figure}
\begin{center}
\begin{tabular}{c}
\includegraphics[width=0.99\textwidth]{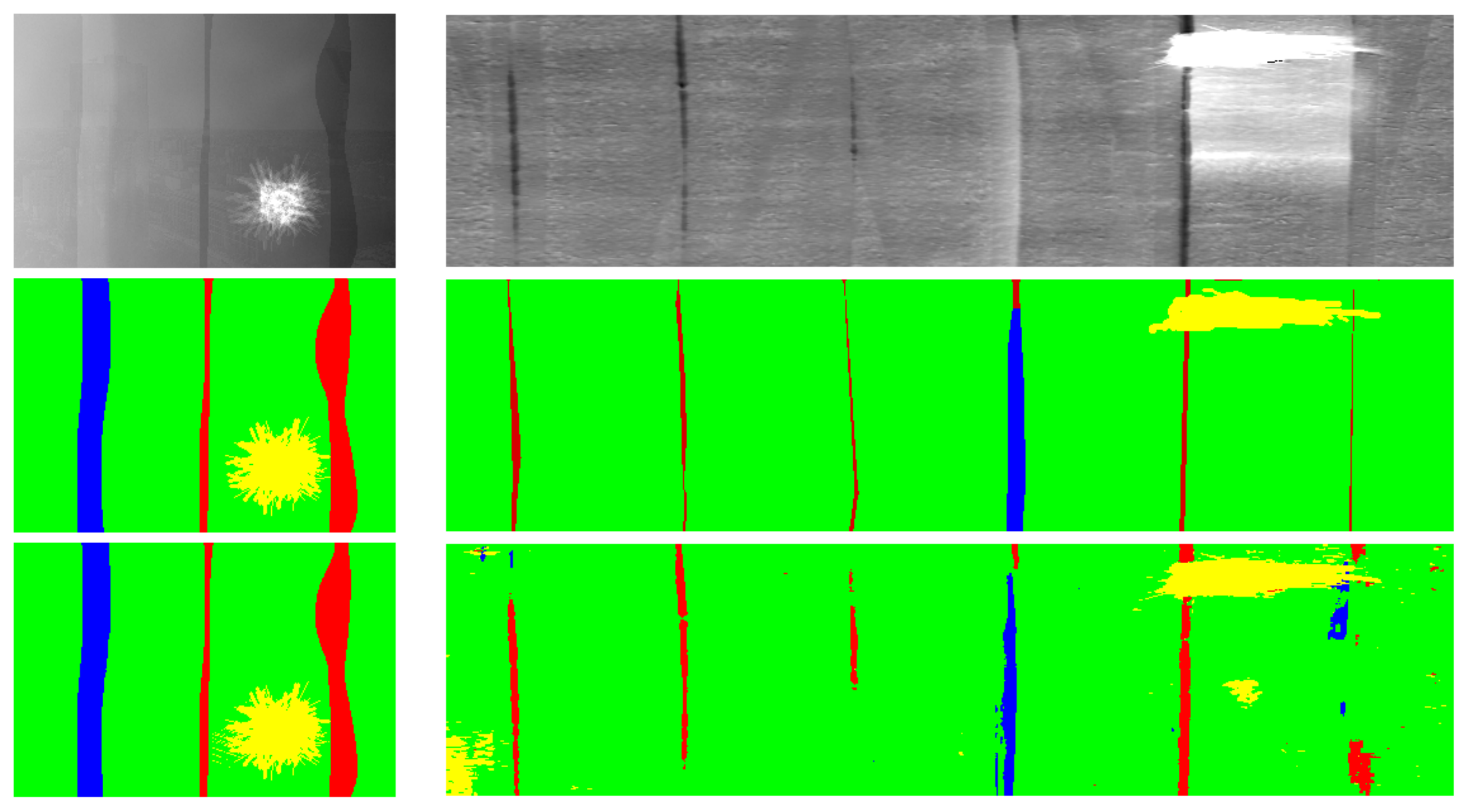}
\end{tabular}
\end{center}
\caption[segResults]
{ 
\label{fig:segResults}: Results of segmentations of unseen artificial data (left) and real data (right). Neural network input (top), ground truth (center), and neural network output (bottom). Pixel labels are visualized with different colors: gap (red), regular tow (green), overlap (blue), and fuzzball (yellow).
}
\end{figure}

We train a neural network on 5000 artificial examples. Each of these examples consists of an artificial depth map and corresponding pixel-wise labels (i.e. ground truth segmentation). After training, the network is validated on a set of 1000 unseen artificial examples and a single real depth map. Input and output size of artificial data is equal to that of training data: 200x300 pixels. The real depth map has a size of 200x800 pixels. Segmentation results are shown in figure \ref{fig:segResults} for artificial (left) and real (right) data. The top row shows (normalized) depth maps which are used as input for the neural network. The center row shows ground truth labels. Ground truth for artificial data is directly derived from hidden variables of the probabilistic model. For the real depth map ground truth comes from manual labeling. The bottom row shows the output of the neural network. 

Details of segmentation performance for validation are shown in tables \ref{tab:confusionArtificial} and \ref{tab:confusionReal}. These tables represent confusion matrices for ground truth and predicted labels in percentage of the total number of pixels. The total number of correctly classified pixels (sum of diagonal elements in the confusion matrix) on average is 99.4\% for 1000 unseen artificial examples and 95.0\% for a real depth map acquired by a real laser triangulation sensor.

All experiments are conducted on a computer with 2x Intel Xeon E5-2650v4 12-Core and NVIDIA Tesla V100 SXM2 32 GB GPU. Total training takes 3 hours and 3 minutes of which artificial data generation consumes 38 minutes. For 100 runs the average duration of a network forward pass takes 6.66ms (standard deviation: 1.04ms) for input depth maps of 200x300 pixels. For a real depth map with size 200x800 pixels, the forward pass takes 15.10ms (standard deviation: 0.72ms).

\begin{table}[h]
%
\caption{Confusion matrix for 1000 unseen artificial test examples. The numbers represent percentages of the total number of pixels.} 
\label{tab:confusionArtificial}
\begin{center}  
\begin{tabular}[t]{ccc|c|c|c|c|}
 & & \multicolumn{4}{c}{\textbf{Ground truth}} \\
\cline{3-7}
{\multirow{5}{*}{\rotatebox[origin=c]{90}{\textbf{Prediction~~~~}}}} & & \multicolumn{1}{|c|}{\textbf{Gap}} & \textbf{Tow} & \textbf{Overlap} & \textbf{Fuzzball} & \textbf{\textbf{$\sum$}} \\ 
\cline{2-7}
 & \multicolumn{1}{|c|}{\textbf{Gap}} & 10.01 & 0.06 & 0.00 & 0.00 & 10.07 \\
\cline{2-7}
 & \multicolumn{1}{|c|}{\textbf{Tow}} & 0.29 & 78.04 & 0.12 & 0.02 & 78.48 \\
\cline{2-7}
 & \multicolumn{1}{|c|}{\textbf{Overlap}} & 0.00 & 0.05 & 6.44 & 0.00 & 6.49  \\
\cline{2-7}
 & \multicolumn{1}{|c|}{\textbf{Fuzzball}} & 0.00 & 0.02 & 0.00 & 4.93 & 4.96 \\
\cline{2-7}
 & \multicolumn{1}{|c|}{\textbf{$\sum$}} & 10.30 & 78.18 & 6.56 & 4.96 & 100.00 \\
\cline{2-7}
\end{tabular}
\end{center}  
\end{table}

\begin{table}[h]
\caption{Confusion matrix for real sensor data. The numbers represent percentages of the total number of pixels.} 
\label{tab:confusionReal}
\begin{center}  
\begin{tabular}[t]{ccc|c|c|c|c|}
 & & \multicolumn{4}{c}{\textbf{Ground truth}} \\
\cline{3-7}
{\multirow{5}{*}{\rotatebox[origin=c]{90}{\textbf{Prediction~~~~}}}} & & \multicolumn{1}{|c|}{\textbf{Gap}} & \textbf{Tow} & \textbf{Overlap} & \textbf{Fuzzball} & \textbf{\textbf{$\sum$}} \\ 
\cline{2-7}
 & \multicolumn{1}{|c|}{\textbf{Gap}} & 1.23 & 1.39 & 0.00 & 0.02 & 2.64 \\
\cline{2-7}
 & \multicolumn{1}{|c|}{\textbf{Tow}} & 0.60 & 91.12 & 0.76 & 0.13 & 92.60 \\
\cline{2-7}
 & \multicolumn{1}{|c|}{\textbf{Overlap}} & 0.00 & 0.64 & 0.62 & 0.01 & 1.27 \\
\cline{2-7}
 & \multicolumn{1}{|c|}{\textbf{Fuzzball}} & 0.01 & 1.41 & 0.00 & 2.07 & 3.49 \\
\cline{2-7}
 & \multicolumn{1}{|c|}{\textbf{$\sum$}} & 1.84 & 94.56 & 1.37 & 2.23 & 100.00 \\
\cline{2-7}
\end{tabular}
\end{center}  
\end{table}

\section{CONCLUSIONS AND FUTURE WORK}

In this paper we propose a probabilistic model to allow a structured approach for the creation of artificial training data. A deep neural network inspired by the U-Net architecture is used to infer pixel labels from observed depth maps. In general, this approach follows the concept of analysis by synthesis. The focus is put on synthesis, i.e. artificial data generating model. The related inference problem is tackled with a deep neural network. We consider this approach appealing because: (1) It enables the use powerful machine learning techniques even if no real data is available. (2) No tedious manual labeling is required. (3) Expert knowledge is directly exploited for the design of the probabilistic model.

Results so far indicate that segmentation quality is lower on real data than on artificial data. At least to some extent this can be explained by the fact that the probabilistic model does not exactly describe the real data generating process. In future work we plan to investigate in more detail how individual parts of the probabilistic model influence segmentation performance. This might help to better understand what are the most important aspects in designing probabilistic models for similar applications.

\acknowledgments     
 
Work presented in this paper has received funding from the European Union’s Horizon 2020 research and innovation programme under grant agreement No 721362 (project “ZAero”) and by the European Union in cooperation with the State of Upper Austria within the project “Investition in Wachstum und Beschäftigung” (IWB).


\bibliography{AV200-68}   
\bibliographystyle{spiebib}   

\end{document}